\documentclass{article} % For LaTeX2e
\usepackage{iclr2023_conference,times}

\usepackage{amsmath,amsfonts,bm}

\def\eqref#1{equation~\ref{#1}}
\def\1{\bm{1}}

\def\vd{{\bm{d}}}

\def\vu{{\bm{u}}}
\def\vv{{\bm{v}}}

\def\vx{{\bm{x}}}

\def\mD{{\bm{D}}}

\def\mI{{\bm{I}}}

\def\mV{{\bm{V}}}

\DeclareMathAlphabet{\mathsfit}{\encodingdefault}{\sfdefault}{m}{sl}
\SetMathAlphabet{\mathsfit}{bold}{\encodingdefault}{\sfdefault}{bx}{n}

\usepackage{hyperref}
\usepackage{url}
\usepackage{graphicx} 
\usepackage{booktabs} 
\usepackage[table]{colortbl} 
\usepackage{algorithm} 
\usepackage{algorithmicx} 
\usepackage{algpseudocode}
\usepackage{caption}
\usepackage{wrapfig}
\usepackage{xcolor}  
\usepackage{mdframed}  
\usepackage{color}
\usepackage{amssymb}

\title{Relay Diffusion: 
Unifying diffusion process across resolutions for image synthesis}

\author{Jiayan Teng$^{\star1}$, Wendi Zheng$^{\star1}$, Ming Ding$^{\star12\dagger}$,\\  \textbf{Wenyi Hong$^1$, Jianqiao Wangni$^2$, Zhuoyi Yang$^1$, Jie Tang$^{1\dagger}$}\\
$^\star$equal contribution\ $^1$Tsinghua University\ $^2$Zhipu AI \ $^\dagger$ corresponding authors\\
\texttt{\{tengjy20@mails,zhengwd23@mails,jietang@mail\}.tsinghua.edu.cn} \\
\texttt{mingding.thu@gmail.com}
} 

\iclrfinalcopy 
\begin{document}
\maketitle
\begin{abstract}
% Diffusion models have become a most promising branch of researches on generative models recently. 
% In despite of this, challenges still remain for diffusion models on handling high-resolution generations, including limited training efficiency as well as brittle setting of noise schedule, etc. 
Diffusion models achieved great success in image synthesis, but still face challenges in high-resolution generation. Through the lens of discrete cosine transformation, we find the main reason is that \emph{the same noise level on a higher resolution results in a higher Signal-to-Noise Ratio in the frequency domain}. In this work, we present Relay Diffusion Model (RDM), which transfers a low-resolution image or noise into an equivalent high-resolution one for diffusion model via blurring diffusion and block noise. Therefore, the diffusion process can continue seamlessly in any new resolution or model without restarting from pure noise or low-resolution conditioning. RDM achieves state-of-the-art FID on CelebA-HQ and sFID on ImageNet 256$\times$256, surpassing previous works such as ADM, LDM and DiT by a large margin. All the codes and checkpoints are open-sourced at \url{https://github.com/THUDM/RelayDiffusion}.
% In this work, we propose Relay Diffusion Model (RDM) as a better framework for diffusion generation. RDM handles image generation with a cascaded pipeline, launching diffusion in each stage from the results of the last stage as the starting point. RDM also introduces adjacency-correlated Gaussian noise, named as block noise, in the process of diffusion, which successfully bridges the gap between generations on different resolutions. We apply RDM on 256$\times$256 image generation and achieve a SoTA FID 3.15 on CelebA-HQ. On the experiment of ImageNet, We achieve a SoTA sFID 3.97 as well as a competitive FID 1.87 (1.79 for SoTA), while only taking 70\% of the training cost of the SoTA model. When classifier-free guidance (CFG) is disabled, RDM outperforms previous SoTA models and already achieves a better FID with only one-third of the training consumption.
% The abstract paragraph should be indented 1/2~inch (3~picas) on both left and
% right-hand margins. Use 10~point type, with a vertical spacing of 11~points.
% The word \textsc{Abstract} must be centered, in small caps, and in point size 12. Two
% line spaces precede the abstract. The abstract must be limited to one
% paragraph.
\end{abstract}

\begin{figure}[H]
    \begin{center}
        {\includegraphics[width=1\linewidth]{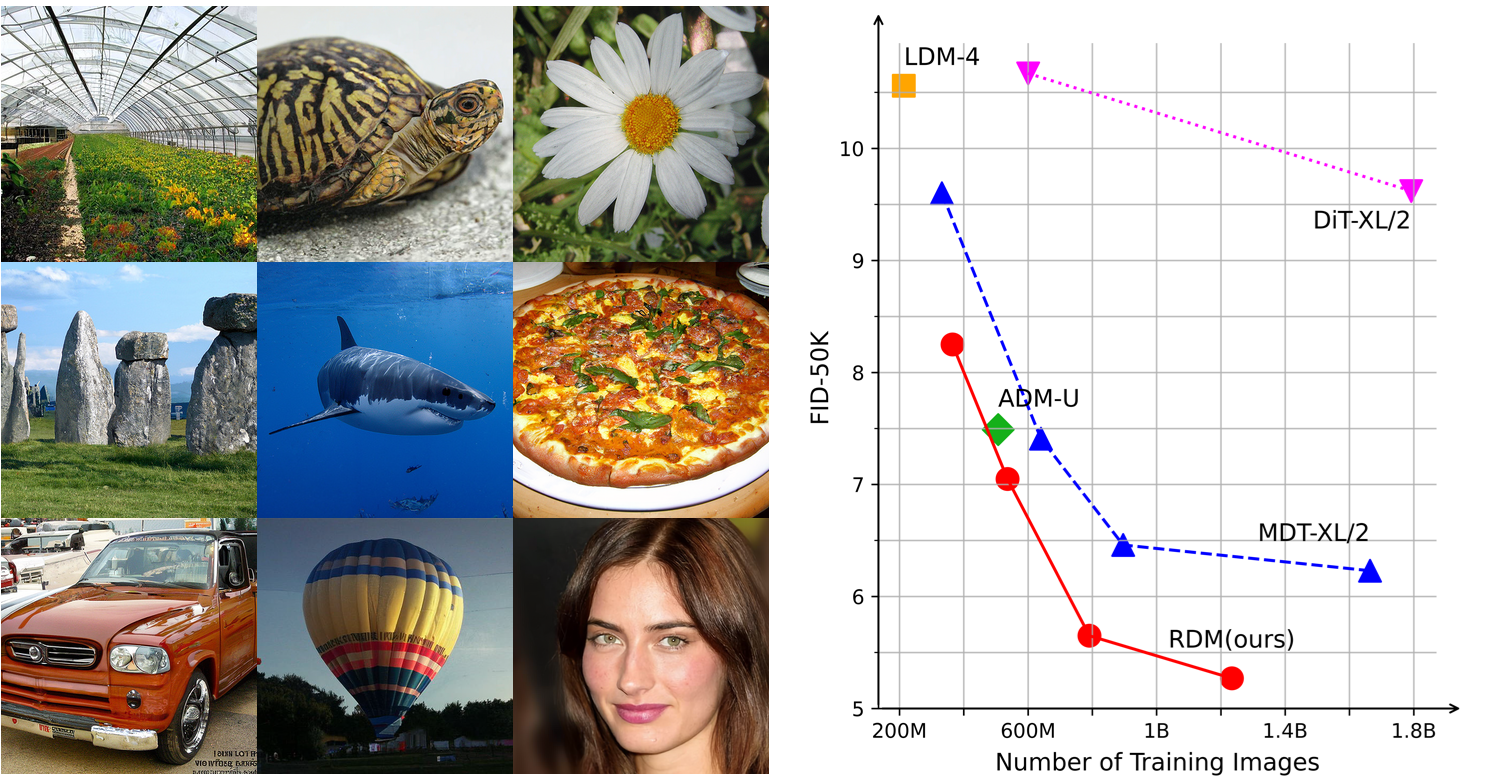}}
    \end{center}
    \caption{(left): Generated Samples by RDM on ImageNet 256$\times$256 and CelebA-HQ 256$\times$256. (right): Benchmarking recent diffusion models on class-conditional ImageNet 256$\times$256 generation without any guidance. RDM can achieve a FID of 1.87 if with classifier-free guidance.}
    \label{fig: main_results}
\end{figure}

\section{Introduction}

Diffusion models~\citep{ho2020denoising,rombach2022high} succeeded GANs~\citep{goodfellow2020generative} and autoregressive models~\citep{ramesh2021zero,ding2021cogview} to become the most prevalent generative models in recent years. However, challenges still exist in the training of diffusion models for high-resolution images. More specifically, there are two main obstacles: 

\textbf{Training Efficiency. } Although equipped with UNet to balance the memory and computation cost across different resolutions, diffusion models still require a large amount of resources to train on high-resolution images. One popular solution is to train the diffusion model on a latent (usually 4$\times$ compression rate in resolution) space and map the result back as pixels~\citep{rombach2022high}, which is fast but inevitably suffers from some low-level artifacts. The cascaded method~\citep{ho2022cascaded, saharia2022photorealistic} trains a series of varying-size super-resolution diffusion models, which is effective but needs a complete sampling for each stage separately.

\textbf{Noise Schedule.} Diffusion models need a noise schedule to control the amount of the isotropic Gaussian noise at each step. The setting of the noise schedule shows great influence over the performance, and most current models follow the linear~\citep{ho2020denoising} or cosine~\citep{nichol2021improved} schedule. However, \emph{an ideal noise schedule should be resolution-dependent} (See Figure~\ref{fig: block} or ~\cite{chen2023importance}), resulting in suboptimal performance to train high-resolution models directly with common schedules designed for resolutions of 32$\times$32 or 64$\times$64 pixels.

These obstacles hindered previous researchers to establish an effective end-to-end diffusion model for high-resolution image generation. \cite{dhariwal2021diffusion} attempted to directly train a 256$\times$256 ADM but found that it performs much worse than the cascaded pipeline. \cite{chen2023importance} and \cite{hoogeboom2023simple} carefully adjusted the hyperparameters of noise schedule and architecture for high-resolution cases, but the quality is still not comparable to the state-of-the-art cascaded methods~\citep{saharia2022photorealistic}. 

In our opinion, the cascaded method contributes in both training efficiency and noise schedule: (1) It provides flexibility to adjust the model size and architecture for each stage to find the most efficient combination. (2) The existence of low-resolution condition makes the early sampling steps easy, so that the common noise schedules (optimized for low-resolution models) can be applied as a feasible baseline to the super-resolution models. Moreover, (3) high-resolution images are more difficult to obtain on the Internet than low-resolution images. The cascaded method can leverage the knowledge from low-resolution samples, meanwhile keep the capability to generate high-resolution images. Therefore, it might not be a promising direction to completely replace the cascaded method with an end-to-end one at the current stage. 

The disadvantages of the cascaded method are also obvious: 
(1) Although the low-resolution part is determined, a complete diffusion model starting from pure noise is still trained and sampled for super-resolution, which is time-consuming. (2) The distribution mismatch between ground-truth and generated low-resolution condition will hurt the performance, so that tricks like conditioning augmentation~\citep{ho2022cascaded} become vitally important to mitigate the gap. Besides, the noise schedule of high-resolution stages are still not well studied.

\textbf{Present Work. } Here we present the \textbf{R}elay \textbf{D}iffusion \textbf{M}odel (RDM), a new cascaded framework to improve the shortcomings of the previous cascaded methods. In each stage, the model starts diffusion from the result of the last stage, instead of conditioning on it and starting from pure noise. Our method is named as the cascaded models work together like a ``relay race''.
The contributions of this paper can be summarized as follows:
\begin{itemize}
    \item We analyze the reasons of the difficulty of noise scheduling in high-resolution diffusion models in frequency domain. Previous works like LDM~\citep{rombach2022high} assume all image signals from the same distribution when analyzing the SNR, neglecting the difference in frequency domain between low-resolution and high-resolution images.
    % However, high-resolution images have a broader frequency spectrum, and the signal strength decreases as the frequency band increases.
    Our analysis successfully account for phenomenon that the same noise level shows different perceptual effects on different resolutions, and introduce the \emph{block noise} to bridge the gap.
    % We propose \emph{block noise}, where the Gaussian noise for spatially adjacent pixels are correlated instead of independent in ordinary diffusion. Block noise mainly corrupts the low-frequency information, increasing the flexibility of the design of noise schedule. This technique not only bridges the gap between different resolutions, but also helps to improve the performance of general diffusion models \zwd{not really}, which is verified by our experiments on CelebA-HQ~\citep{karras2018progressive}. \ming{todo rewrite}
    \item We propose RDM to disentangle the diffusion process and the underlying neural networks in the cascaded pipeline. RDM gets rid of the low-resolution conditioning and its distribution mismatch problem. Since RDM starts diffusion from the low-resolution result instead of pure noise, the training and sampling steps can also be reduced. 
    \item We evaluate the effectiveness of RDM on unconditional CelebA-HQ 256$\times$256 and conditional ImageNet 256$\times$256 datasets. RDM achieves state-of-the-art FID on CelebA-HQ and sFID on ImageNet.  
\end{itemize}

\section{Preliminary}

\subsection{Diffusion Models}
To model the data distribution $p_{data}(\mathbf{x}_0)$, denoising diffusion probabilistic models (DDPMs,~\citet{ho2020denoising}) define the generation process as a Markov chain of learned Gaussian transitions. DDPMs first assume a forward diffusion process, corrupting real data $\mathbf{x}_0$ by progressively adding Gaussian noise from time steps $0$ to $T$, whose variance $\{\beta_t\}$ is called the noise schedule:
\begin{equation}
    q(\mathbf{x}_t|\mathbf{x}_{t-1})=\mathcal{N}(\mathbf{x}_t; \sqrt{1-\beta_t}\mathbf{x}_{t-1},\beta_t\mathbf{I}).
\end{equation}
The reverse diffusion process is learned by a time-dependent neural network to predict denoised results at each time step, by optimizing the variational lower bound (ELBO).

Many other formulations for diffusion models include stochastic differential equations (SDE,~\citet{song2020score}), denoising diffusion implicit models (DDIM,~\citet{song2020denoising}), etc. \cite{karras2022elucidating} summaries these different formulations into the \textbf{EDM} framework.
% EDM~\citep{karras2022elucidating} achieves significant improvement of generation quality by introducing new noise scheduling, loss weighting and sampler discretization.
In this paper, we generally follow the EDM formulation and implementation.
The training objective of EDM is defined as $L_2$ error terms:
\begin{equation}
    \mathbb{E}_{\mathbf{x}\sim p_{data},\sigma\sim p(\sigma)}\mathbb{E}_{\mathbf{\epsilon}\sim\mathcal{N}(\mathbf{0}, \mathbf{I})}\Vert D(\mathbf{x}+\sigma\mathbf{\epsilon},\sigma)-\mathbf{x}\Vert^2,
\end{equation}
where $p(\sigma)$ represents the distribution of a continuous noise schedule. $D(\mathbf{x}+\mathbf{\epsilon},\sigma)$ represents the denoiser function depending on the noise scale. We also follow the EDM precondition for $D(\mathbf{x}+\mathbf{\epsilon},\sigma)$ with $\sigma$-dependent skip connection~\citep{karras2022elucidating}.

% It often requires a careful tuning of the noise schedule to implement diffusion models on high-resolution generation~\citep{chen2023importance}. \cite{ho2022cascaded} instead proposes Cascaded Diffusion Models (CDM), which divides high-resolution generation into multiple stages. At the first stage, CDM generates low-resolution images. After that, CDM iteratively performs super-resolution generation conditioning on the outputs of the previous stage. Similar designs are also extensively adopted in general-domain text-to-image generation researches including GLIDE~\citep{nichol2021glide}, Imagen~\citep{saharia2022photorealistic}, DALL-E~\citep{ramesh2022hierarchical} and eDiff-I~\citep{balaji2022ediffi}.

Cascaded diffusion model (CDM,~\citet{ho2022cascaded}) is proposed for high-resolution generation. CDM divides the generation into multiple stages, where the first stage generates low-resolution images and the following stages perform super-resolution conditioning on the outputs of the previous stage. Cascaded models are extensively adopted in recent works of text-to-image generation, e.g. Imagen~\citep{saharia2022photorealistic}, DALL-E-2~\citep{ramesh2022hierarchical} and eDiff-I~\citep{balaji2022ediffi}.

\subsection{Blurring Diffusion} \label{sec:ihd}

The Inverse Heat Dissipation Model (IHDM)~\citep{rissanen2022generative} generates images by reversing the heat dissipation process. The heat dissipation is a thermodynamic process describing how the temperature $u(x, y, t)$ at location $(x,y)$ changes in a (2D) space with respect to the time $t$. The dynamics can be denoted by a PDE $\frac{\partial u}{\partial t}= \frac{\partial^2 u}{\partial x^2}+\frac{\partial^2 u}{\partial y^2}$.

% Blurring diffusion models~\citep{rissanen2022generative,hoogeboom2022blurring} generates images by reversing the heat dissipation process. \jq{The heat dissipation, or heat transfer}  is a physical \jq{thermodynamic} process describing how the temperature $u(x, y, t)$ \jq{at location $(x,y)$} changes in a (2D) space \jq{w.r.t. time $t$}. The dynamics can be denoted by a PDE $\frac{\partial u}{\partial t}= \jq{\propto} \frac{\partial^2}{\partial x^2}+\frac{\partial^2}{\partial y^2}$, \jq{here the linear coefficient is related to the thermal conductivity of the object, which we assume to be homogeneous isotropic material and have a uniform constant}.

% If the temperature $\mathcal{D}$ is a function of time $t$ and the coordinate $(x,y)$, we can write the partial differential equation (PDE) for heat dissipation as , where $\Delta=\frac{\partial^2}{\partial x^2}+\frac{\partial^2}{\partial y^2}$ denotes the laplace operator.
% In contrast to the standard diffusion, heat dissipation as well defines a forward process of data destruction that smooths the internal difference.

Blurring diffusion~\citep{hoogeboom2022blurring} is further derived by augmenting the Gaussian noise with heat dissipation for image corruption. Since simulating the heat equation up to time $t$ is equivalent to a convolution with a Gaussian
kernel with variance $\sigma^2=2t$ in an infinite plane~\citep{bredies2018mathematical}, the intermediate states $\vx_t$ become blurry, instead of noisy in the standard diffusion. If Neumann boundary conditions are assumed, blurring diffusion in discrete 2D pixel space can be transformed to the frequency space by Discrete Cosine Transformation (DCT) conveniently as:
\begin{equation}
    q(\vu_t | \vu_0) = \mathcal{N}(\vu_t | \mD_t \vu_{0}, \sigma_t^2\mI),
    \label{eq:blur-diff}
\end{equation}
where $\vu_t = \text{DCT}(\vx_t)$
% \tjy{$=\mV^\mathrm{T}\vx_t$}
, and $\mD_t=e^{\mathbf{\Lambda} t}$ is a diagonal matrix with $\mathbf{\Lambda}_{i\times W+j}=-\pi^2(\frac{i^2}{H^2}+\frac{j^2}{W^2})$ for coordinate $(i,j)$. Here Gaussian noise with variance $\sigma_t^2$ is mixed into the blurring diffusion process to transform the deterministic dissipation process to a stochastic one for diverse generation~\citep{rissanen2022generative}.

\section{method}

% structure setting
% block noise
% cascade model
% stochastic sampler
\begin{figure}[h]
    \begin{center}
    {\includegraphics[width=1\linewidth]{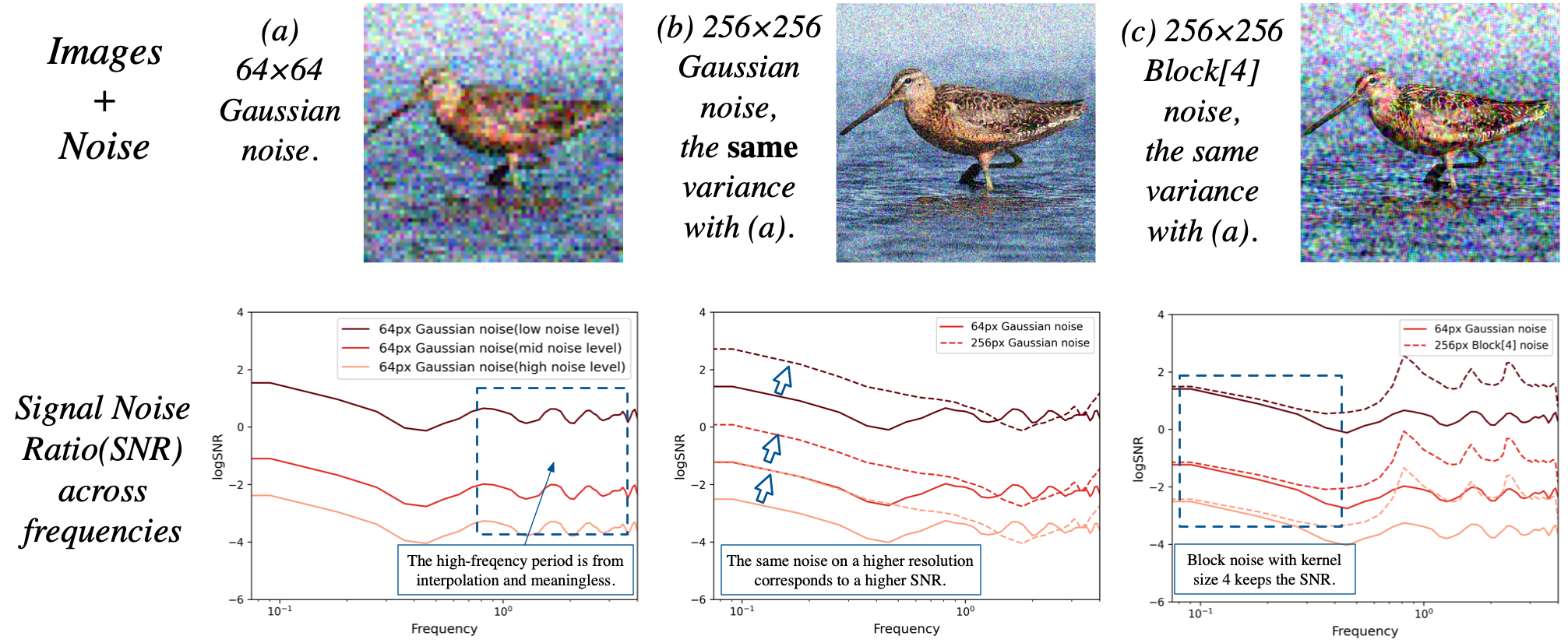}}
    \end{center}
    \caption{Illustration of spatial and frequency results after adding independent Gaussian and block noise. (a)(b) At the resolution of $64\times 64$ and $256\times 256$, the same noise level results in different perceptual effects, and in the frequency plot, the SNR curve shifts upward. (c) The independent Gaussian noise at the resolution $64\times 64$ and block noise (kernel size = 4)  at the resolution $256\times 256$ produces similar results in both spatial domain and frequency domain. The noise is $\mathcal{N}(0,0.3^2)$ for (a). These SNR curves are universally applicable to most natural images.
    % , and see Appendix~\ref{app:freq} for details.
    }
    % \caption{ Adding noise of the same level to images ($\vx_t = \vx_0 + \sigma \bm{\epsilon}$ with $\sigma = 0.3$) of different resolutions. For block noise, we add it with $s = resolution/32$.} %是否需要像这样说明我们在图中是怎么加的block noise
    \label{fig: block}
    \end{figure}

\subsection{Motivation}\label{sec:motivation}
The noise schedule is vitally important to the diffusion models and resolution-dependent. A certain noise level appropriately corrupting the $64\times 64$ images, could fail to corrupt the $256\times 256$ (or a higher resolution) images, which is shown in the first row of Figure~\ref{fig: block}(a)(b). \cite{chen2023importance} and \cite{hoogeboom2023simple} attributed this to the lack of schedule-tuning, but we found an analysis from the perspective of frequency spectrum can help us better understand this phenomenon.

\textbf{Frequency spectrum analysis of the diffusion process.} The natural images with different resolutions can be viewed as the result of visual signals sampled at varying frequencies. To compare the frequency features of a $64\times 64$ image and a $256\times 256$ image, we can upsample the $64\times 64$ one to $256\times 256$, perform DCT and compare them in the 256-point DCT spectrum.
% In frequency domain, the power spectrum density (PSD) of a natural image is a 
% As the resolution increases, adjacent pixels are more likely to be similar, leading to a decrease in power spectral density with increasing frequency in the frequency spectrum analysis DCT.
% The noise in diffusion models in the frequency domain manifests as a horizontal line, determined by the variance. 
The second row of Figure~\ref{fig: block}(a) shows the signal noise ratio (SNR) at different frequencies and diffusion steps. In Figure~\ref{fig: block}(b), we clearly find that \emph{the same noise level on a higher resolution results in a higher SNR in the (low-frequency part of) the frequency domain}. 
% The root cause of this phenomenon stems from the characteristic that \emph{the higher the resolution of the image, the closer the adjacent pixels are}. 
Detailed frequency spectrum analysis are included in Appendix~\ref{app:psd}.

At a certain diffusion step, a higher SNR means that during training the neural network presumes the input image more accurate, but 
the early steps may not be able to generate such accurate images after the increase in SNR. This training-inference mismatch will accumulate over step by step during sampling, leading to the degradation of performance. 

\textbf{Block noise as the equivalence at high resolution.} After the upsampling from $64\times 64$ to $256\times 256$, the independent Gaussian noise on $64\times 64$ becomes noise on $4\times 4$ grids, thus greatly changes its frequency representation.
To find a variant of the $s\times s$-grid noise without deterministic boundaries, we propose \textbf{Block noise}, where the Gaussian noise are correlated for nearby positions. More specifically, the covariance between noise $\epsilon_{x_0,y_0}$ and $\epsilon_{x_1, y_1}$ is defined as
\begin{equation}
    \text{Cov}(\epsilon_{x_0,y_0}, \epsilon_{x_1, y_1}) = \frac{\sigma^2}{s^2}\max\big(0, s-\text{dis}(x_0,x_1)\big)\max\big(0, s-\text{dis}(y_0,y_1)\big),
\end{equation}
where $\sigma^2$ is the noise variance, and $s$ is a hyperparameter \emph{kernel size}. The dis$(\cdot, \cdot)$ function here is the Manhattan distance. For simplicity, we ``connect'' the top and bottom edges and the left and right edges of the image, resulting in
\begin{equation}
    \text{dis}(x_0,x_1) = \min\left( |x_0 - x_1|, x_{max} - |x_0 - x_1| \right).
\end{equation}
The block noise with kernel size $s$ can be generated by averaging $s\times s$ independent Gaussian noise. Suppose we have an independent Gaussian noise matrix $\epsilon$, the block noise construction function Block$[s](\cdot)$ is defined as
\begin{equation}
	\text{Block}[s](\epsilon)_{x,y}= \frac{1}{s}\sum_{i=0}^{s-1} \sum_{j=0}^{s-1}\epsilon_{x-i,y-j},
    \label{eq:block-noise}
\end{equation}
where $\text{Block}[s](\epsilon)_{x,y}$ is the block noise at the position $(x,y)$, and $\epsilon_{-x} = \epsilon_{x_{max}-x}$. Figure~\ref{fig: block}(c) shows that the block noise with kernel size $s=4$ on $256\times 256$ has a similar frequency spectrum as the independent Gaussian noise on $64\times 64$ images.

The analysis above seems to indicate that we can design an end-to-end model for high-resolution images by introducing block noise in early diffusion steps, while cascaded models already achieves great success. Therefore, a revisit of the cascaded models is necessary.

\textbf{Why does the cascaded models alleviate this issue?} Experiments in previous works~\citep{nichol2021improved,dhariwal2021diffusion} have already shown that cascaded models perform better than end-to-end models under a fair setting. These models usually use the same noise schedule in all stages, so why are the cascaded models not affected by the increase of SNR? The reason is that in the super-resolution stages, the low-resolution condition greatly ease the difficulty of the early steps, so that even the higher SNR requires a more accurate input, the accuracy is within the capability of the model.

A natural idea is that since the low-frequency information in the high-resolution stage has already been determined by the low-resolution condition, we can continue generating directly from the upsampled result to reduce both the training and sampling steps. 
However, the generation of low-resolution images is not perfect, and thus the solution of the distribution mismatch between ground-truth
and generated low-resolution images is a priority to ``continue'' the diffusion process.

\subsection{Relay Diffusion}\label{sec:framework}

\begin{figure}[h]
\begin{center}
{\includegraphics[width=1\linewidth]{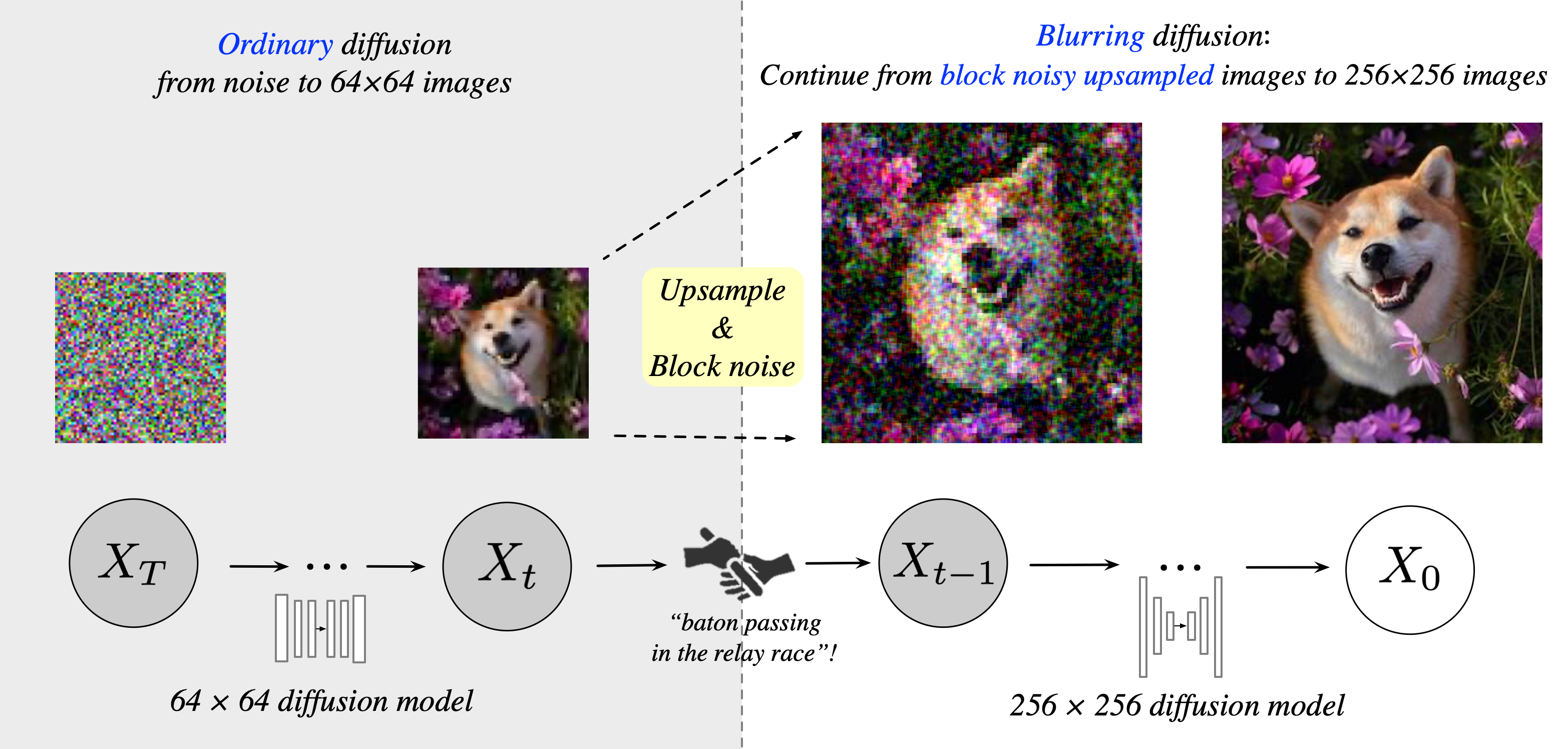}}
\end{center}
\caption{ Pipeline of Relay Diffusion Models (RDM). }
\label{fig:pipeline}
\end{figure}
We propose relay diffusion model (RDM), a cascaded pipeline connecting the stages with block noise and (patch-level) blurring diffusion. Different from CDM, RDM considers the equivalence of the low-resolution generated images when upsampled to high resolution. Suppose that the generated $64\times 64$ low-resolution image $\mathbf{x}^L_0 = \mathbf{x}^L + \epsilon_L$ can be decomposed into a sample in real distribution $\mathbf{x}^L$ and a remaining noise $\epsilon_L \sim \mathcal{N}(\mathbf{0}, \beta_0^2\mathbf{I})$. As mentioned in section~\ref{sec:motivation}, the $256\times 256$ equivalence of $\epsilon_L$ is Block[$4$] noise with variance $\beta_0^2$, denoted by $\epsilon_H$. After (nearest) upsampling, $\mathbf{x}^L$ becomes $\mathbf{x}^H$, where each $4\times 4$ grid share the same pixel values. We can define it as the starting state of a \emph{patch-wise blurring diffusion}.

% Distinct from formerly proposed frameworks of cascaded diffusion models that conditions super-resolution on previous stage's outputs, RDM combines two stages of generation in a unified diffusion schedule \jq{(distinct from a 'seperate' framework, RDM is a unified ...)}. The first stage of RDM generates images at a low resolution, e.g. 64$\times$64, with ordinary diffusion modeling. RDM then upsamples the first stage's generation into blurry images at a higher resolution, e.g. 256$\times$256, to be the starting point of the super-resolution stage \jq{deblurring/ inverse heat dissipation}, \jq{and then repeat this procedure to convergence,} just the same as a ``baton pass in the relay race".

% In the super-resolution \jq{deblurring/ inverse heat dissipation} stage of RDM, we combine the forward process of ordinary diffusion with the heat dissipation from~\cite{rissanen2022generative}, for the adaptation of the blurry starting point. 
Unlike blurring diffusion models~\citep{rissanen2022generative}~\citep{hoogeboom2022blurring} that perform the heat dissipation on the entire space of images, we propose to implement the heat dissipation on each $4\times 4$ patch independently, which is of the same size as the upsampling scale.
% We name such a forward blurring process as \emph{patch-wise heat dissipation}.
We first define a series of patch-wise blurring matrix $\{\mD^p_t\}$, which is introduced in detail in Appendix~\ref{sec:patch_blur}. The forward process would have a similar representation with~\eqref{eq:blur-diff}:
\begin{equation}
    q(\vx_t | \vx_0) = \mathcal{N}(\vx_t | \mV\mD^p_t\mV^\mathrm{T}\vx_{0}, {\sigma_t}^2\mI), \quad t\in\{0,..,T\},\label{eq:patch-blur}
\end{equation}
where $\mV^\mathrm{T}$ is the projection matrix of DCT and $\sigma_t$ is the variance of noise. Here the $\mD^p_T$ is chosen to guarantee $\mV\mD^p_T\mV^\mathrm{T}\vx_{0}$ in the same distribution as $\vx^H$, meaning that the blurring process ultimately makes the pixel value in each $4\times 4$ patch the same.

The training objective of the high-resolution stage of RDM generally follows EDM~\citep{karras2022elucidating} framework in our implementation. The loss function is defined on the prediction of denoiser function $D$ to fit with true data $\vx$, which is written as:
\begin{align}
    &\mathbb{E}_{\vx\sim p_{data},t\sim \mathcal{U}(0,1),\mathbf{\epsilon}\sim\mathcal{N}(\mathbf{0},\mathbf{I})}\Vert D(\vx_t,\sigma_t)-\vx\Vert^2, \nonumber\\
    &\text{where}\quad \vx_t=\textcolor{blue}{\underbrace{\mV\mD^p_t\mV^\mathrm{T}}_{blurring}}\vx+\frac{\sigma}{\sqrt{1+\alpha^2}}\big(\mathbf{\epsilon}+\alpha\cdot\textcolor{red}{\underbrace{\text{Block}[s](\mathbf{\epsilon'})}_{block\ noise}}\big),\label{eq:relaydiff}
\end{align}
where $\epsilon$ and $\epsilon'$ are two independent Gaussian noise.
The main difference in training between RDM and EDM is that the corrupted sample $\vx_t$ is not simply $\vx_t=\vx + \epsilon$, but a mixture of the blurred image, block noise and independent Gaussian noise. Ideally, the noise should gradually transfer from block noise to high-resolution independent Gaussian noise, but we find that a weighting average strategy perform well enough, because the low-frequency component of the block noise is much larger than the independent Gaussian noise, and vice versa for high-frequency component. $\alpha$ is a hyperparameter and the normalizer $\frac{1}{\sqrt{1+\alpha^2}}$ is used to keep the variance of the noise, $\sigma^2$ unchanged.

The advantages of RDM compared to CDM includes:
\begin{itemize}
    \item RDM is more efficient, because RDM skips the re-generation of low-frequency information in the high-resolution stages, and reduce the number of training and sampling steps.
    \item RDM is more simple, because it gets rid of the low-resolution conditioning and conditioning
    augmentation tricks. The consumption from cross-attention with the low-resolution condition is also spared.
    \item RDM is more potential in performance, because RDM is a Markovian denosing process (if with DDPM sampler). Any artifacts in the low-resolution images can be corrected in the high-resolution stage, while CDM is trained to correspond to the low-resolution condition.
\end{itemize}
Compared to end-to-end models~\citep{chen2023importance,hoogeboom2023simple},
\begin{itemize}
    \item RDM is more flexible to adjust the model size and leverage more low-resolution data.
\end{itemize} 

\subsection{Stochastic Sampler} \label{sec:sampler}
Since RDM differs from traditional diffusion models in the forward process, we also need to adapt the sampling algorithms. In this section, we focus on the EDM sampler~\citep{karras2022elucidating} due to its flexibility to switch between the first and second order (Heun's) samplers.

% For the super-resolution stage of RDM, we follow EDM~\citep{karras2022elucidating} to a Heun's second-order sampler for the generation. 

Heun's method introduces an additional step for the correction of the first-order sampling. The updating direction of a first-order sampling step is controlled by the gradient term $\vd_n=\frac{\vx_n-\vx_\theta(\vx_n,\sigma_{t_n})}{\sigma_{t_n}}$. The correction step updates current states with an averaged gradient term $\frac{\vd_n+\vd_{n-1}}{2}$. Heun's method takes account of the change of gradient term $\frac{\vd\vx}{\vd t}$ between $t_n$ and $t_{n-1}$. Therefore, it achieves higher quality while allowing for fewer steps of sampling.

We adapt the EDM sampler to the blurring diffusion of RDM's super-resolution stage following the derivation of DDIM~\citep{song2020denoising}. We define the indices of sampling steps as $\{t_i\}_{i=0}^N$, in corresponding to the noisy states of images $\{\vx_i\}_{i=0}^N$. To apply blurring diffusion, images are transformed into frequency space by DCT as $\vu_i=\mV^\mathrm{T}\vx_i$. \cite{song2020denoising} uses a family of inference distributions to describe the diffusion process. We can write it for blurring diffusion as:
\begin{equation} \label{eq:family_dist}
    q_{\delta}(\vu_{1:N}|\vu_0)=q_{\delta}(\vu_N|\vu_0)\prod\limits_{n=2}^N q_{\delta}(\vu_{n-1}|\vu_n,\vu_0),
\end{equation}
where $\delta\in \mathbb{R}^\mathrm{N}_{\geq 0}$ denotes the index vector for the distribution. For all $n>1$, the backward process is:
\begin{equation}\label{eq:sampler_dist}
    q_{\delta}(\vu_{n-1}|\vu_n,\vu_0)=\mathcal{N} \big(\vu_{n-1}|\frac{1}{\sigma_{t_n}}(\sqrt{\sigma^2_{t_{n-1}}-\delta^2_n}\vu_n+(\sigma_{t_n}\mD^p_{t_{n-1}}-\sqrt{\sigma^2_{t_{n-1}}-\delta^2_n}\mD^p_{t_n})\vu_0),\delta^2_n\mI \big).
\end{equation}
The mean of the normal distribution ensures the forward process to be consistent with the formulation of blurring diffusion in Section~\ref{sec:framework}, which is $q(\vu_n | \vu_0) = \mathcal{N}(\vu_n | \mD^p_{t_n} \vu_{0},\sigma_{t_n}^2\mI)$. When the index vector $\delta$ is $0$, the sampler degenerates into an ODE sampler. We set $\delta_n=\eta\sigma_{t_{n-1}}$ for our sampler, where $\eta\in[0,1)$ is a fixed scalar controlling the scale of randomness injected during sampling. We substitute the definition into Eq.~\ref{eq:sampler_dist} to obtain our sampler function as:
% \begin{equation}\label{eq:sampler}
%     \vu_{n-1}=\frac{\mD^p_{t_{n-1}}}{\mD^p_{t_n}}\vu_n+(\sqrt{1-\eta^2}\frac{\sigma_{t_{n-1}}}{\sigma_{t_n}}-\frac{\mD^p_{t_{n-1}}}{\mD^p_{t_n}})\frac{\vu_n-\mD^p_{t_n}\Tilde{\vu}_0}{\sigma_{t_n}}+\eta\sigma_{t_{n-1}}\bm{\epsilon}.
% \end{equation}
\begin{equation}\label{eq:sampler}
    \vu_{n-1}=(\mD^p_{t_{n-1}} + \gamma_n(\mI - \mD^p_{t_n}))\vu_n + \sigma_{t_n}(\gamma_n\mD^p_{t_n} - \mD^p_{t_{n-1}})\frac{\vu_n-\Tilde{\vu}_0}{\sigma_{t_n}} + \eta\sigma_{t_{n-1}}\bm{\epsilon},
\end{equation}
where $\gamma_n\triangleq\sqrt{1-\eta^2}\frac{\sigma_{t_{n-1}}}{\sigma_{t_n}}$. As in the section~\ref{sec:motivation}, we also need to consider block noise besides blurring diffusion. The adaptation is just to replace isotropic Gaussian noise $\bm{\epsilon}$ with $\Tilde{\bm{\epsilon}}$, which is a weighted sum of the block noise and isotropic Gaussian noise. $\Tilde{\vu}_0=\vu_\theta(\vu_n,\sigma_{t_n})$ is predicted by the neural network.

Finally, a stochastic sampler for the super-resolution stage of RDM is summaries in Algorithm~\ref{alg:sample}. We provide a detailed proof of the consistency between our sampler and the formulation of blurring diffusion in Appendix~\ref{sec:sampler_deriv}.

\begin{algorithm}
\caption{the RDM second-order stochastic sampler}
\begin{algorithmic}
\State {{\bf sample} $\vx_N \sim \mathcal{N} \big( \mathbf{0}, \sigma_N^2 \mathbf{I} \big)$}
\State $\vu_N=\mV^\mathrm{T}\vx_N$
    \textcolor{gray}{\Comment{transformed into the frequency domain}}
\For{$n \in \{N, \dots, 1\}$}
    \State $\gamma_n=\sqrt{1-\eta^2}\frac{\sigma_{t_{n-1}}}{\sigma_{t_n}},\quad \delta_n=\eta\sigma_{t_{n-1}}$
        \textcolor{gray}{\Comment{coefficient of the random term}}
    \State $\Tilde{\vu}_0=\vu_\theta(\vu_n, \sigma_{t_n})$
        \textcolor{gray}{\Comment{model prediction at $t_n$}}
    \State {$\vd_n=\frac{\vu_n - \Tilde{\vu}_0}{\sigma_{t_n}}$} 
        \textcolor{gray}{\Comment{first-order gradient term at $t_n$}}
    \State $\vu_{n-1}=(\mD^p_{t_{n-1}} + \gamma_n(\mI - \mD^p_{t_n}))\vu_n + \sigma_{t_n}(\gamma_n\mD^p_{t_n} - \mD^p_{t_{n-1}})\vd_n + \delta_n\bm{\epsilon}$
    \State $ $
        \textcolor{gray}{\Comment{from $t_n$ to $t_{n-1}$ using Euler's method}}

    \begin{mdframed}[backgroundcolor=yellow!10,hidealllines=true,innerleftmargin=0pt,innerrightmargin=0pt]
    \If{$n\ne 1$} \textcolor{gray}{\Comment{the second-order part}}
        \State $\Tilde{\vu}'_0=\vu_\theta(\vu_{n-1},\sigma_{t_{n-1}})$
            \textcolor{gray}{\Comment{model prediction at $t_{n-1}$}}
        \State $\vd_{n-1}=\frac{\vu_{n-1}-\Tilde{\vu}'_0}{\sigma_{t_{n-1}}}$
            \textcolor{gray}{\Comment{gradient term at $t_{n-1}$}}
        \State $\vd'_n=\frac{\vd_n+\vd_{n-1}}{2}$
            \textcolor{gray}{\Comment{second-order gradient term}}
        \State $\vu'_{n-1}=(\mD^p_{t_{n-1}} + \gamma_n(\mI - \mD^p_{t_n}))\vu_n + \sigma_{t_n}(\gamma_n\mD^p_{t_n} - \mD^p_{t_{n-1}})\vd'_n + \delta_n\bm{\epsilon}$ \textcolor{gray}{\Comment{correction}}            
    \EndIf
    \end{mdframed}
    \State $\vu_{n-1}=\vu'_{n-1}$
\EndFor
\State $\vx_0=\mV\vu_0$
\end{algorithmic}
\label{alg:sample}
\end{algorithm}

% \begin{algorithm}
% \caption{$2^{\text{nd}}$ order stochastic sampler}
% \begin{algorithmic}
% \State {{\bf sample} $\vx_N \sim \mathcal{N} \big( \mathbf{0}, \sigma_{max}^2 \mathbf{I} \big)$}
% \For{$t \in \{N, \dots, 1\}$}
%     \State $g_t=\eta \sigma_{t-1},\ \gamma_t=\frac{\sqrt{\sigma_{t-1}^2-g_t^2}}{\sigma_t}$
%     % \State $a_t=\frac{\sqrt{\sigma_{t-1}^2-g_t^2}}{\sigma_t}$
%     % \State $b_t=\bm{\alpha}_{t-1}- a_t \bm{\alpha}_t$
%     % \State $\vx_{t-1}=a_t \vx_t+b_t \vx_\theta(\vx_t,\sigma_t)+g_t \bm{\hat{\epsilon}}$
%     \State {$\vd_t=\frac{\vx_t - \vx_\theta( \vx_t, \sigma_t)}{\sigma_t}$} 
%         \textcolor{gray}{\Comment{construct gradient term}}
%     \State $\vx_{t-1}= \mV ((\bm{\alpha}_{t-1} + \gamma_t(\bm{I}-\bm{\alpha}_t)) \mV^{\mathrm{T}} \vx_t
%     + \sigma_t(\gamma_t \bm{\alpha}_t - \bm{\alpha}_{t-1}) \mV^{\mathrm{T}} \vd_t)
%     + g_t \Tilde{\bm{\epsilon}}
%     $
%     \If{$\sigma_{t-1} \ne 0$}
%         \State {$\vd_t^{'}=\frac{\vx_{t-1} - \vx_\theta( \vx_{t-1}, \sigma_{t-1})}{\sigma_{t-1}}$}
%             \textcolor{gray}{\Comment{$2^{\text{nd}}$-order}}
%         \State $\vx_{t-1}= \mV ((\bm{\alpha}_{t-1} + \gamma_t(\bm{I}-\bm{\alpha}_t)) \mV^{\mathrm{T}} \vx_t
%         + \sigma_t( \gamma_t \bm{\alpha}_t - \bm{\alpha}_{t-1}) \mV^{ \mathrm{T}} \frac{ \vd_t+ \vd_t^{'}}{2})
%         + g_t \Tilde{\bm{\epsilon}}
%         $
%     \EndIf
% \EndFor
% \end{algorithmic}
% \label{alg:sample}
% \end{algorithm}
\section{Experiments}

\subsection{Experimental Setting}

\textbf{Dataset.} We use CelebA-HQ and ImageNet in our experiments. CelebA-HQ~\citep{karras2018progressive} is a high-quality subset of CelebA~\citep{liu2015deep} which consists of 30,000 images of faces from human celebrities. ImageNet~\citep{deng2009imagenet} contains 1,281,167 images spanning 1000 different classes and is a widely-used dataset for generation and vision tasks. We train RDM on these datasets to generate $256\times 256$ images.

% \textbf{Dataset.} We train RDM on datasets including unconditional CelebA-HQ $256\times 256$ and conditional ImageNet $256\times 256$. ImageNet~\citep{deng2009imagenet} is a widely used conditional dataset for image generation, image recognition and so on. We choose the training set of ILSVRC, the most frequently used subset of ImageNet, for our experiments. It contains 1,281,167 images spanning 1000 different classes. 
% CelebA-HQ~\citep{karras2018progressive} is a high-quality version of CelebA. it consists of 30,000 images of diverse human faces.

\textbf{Architecture and Training.} RDM adopts UNet~\citep{ronneberger2015u} as the backbone of diffusion models for both the first and the second stage. The detailed architectures largely follow ADM~\citep{dhariwal2021diffusion} for fair comparison. We train unconditional models on CelebA-HQ and class-conditional models on ImageNet respectively. Since we follow the EDM implementation, we directly use the released checkpoint from EDM in ImageNet in the $64\times 64$ stage. The FLOPs of the $64\times 64$ model are about $1/10$ that of the $256\times 256$ model.
% For the two tasks above, we calculate the overall consumption of training or sampling by summing up the steps from both two stages, while dividing steps from the first stage by ten for a general equivalence on GFLOPs (104 GFLOPs for first stage and 1,117GFLOPs for second stage).
See Appendix~\ref{app:hypers} for more information about the architecture and hyperparameters of RDM.

% \textbf{Architecture and Training.} RDM follow~\citet{dhariwal2021diffusion} using the ADM architecture which utilize UNet as the backbone in both the first and the second stage. We train RDM on conditional ImageNet and unconditional CelebA-HQ dataset separately. 
% The first stage of RDM with 104 GFLOPs has 295 million parameters, while the second stage of RDM with 1117 GFLOPs has 553 million parameters. 
% For fairness, the training steps of the first stage was divided by 10 to be equivalent to the training steps of the second stage based on their GFLOPs. See Table \ref{tbl:hypers} for detailed architecture and hyperparameters of RDM.

\textbf{Evaluation.} We use metrics including FID~\citep{heusel2017gans}, sFID~\citep{nash2021generating}, IS~\citep{salimans2016improved}, Precision and Recall~\citep{kynkaanniemi2019improved} for a comprehensive evaluation of the results. FID measures the difference between the features of model generations and real images, which is extracted by a pretrained Inception network. sFID differs from FID by using intermediate features, which better measures the similarity of spatial distribution. IS and Precision both measure the fidelity of the samples, while Recall indicates the diversity. We compute metrics with 50,000 and 30,000 generated samples for ImageNet and CelebA-HQ respectively.

% \textbf{Evaluation.} Our evaluation comprehensively considers both the quality and diversity of the generated images, encompassing metrics such as FID~\citep{heusel2017gans}, sFID~\citep{nash2021generating}, IS~\citep{salimans2016improved}, Precision and Recall~\citep{kynkaanniemi2019improved}. 
% FID utilizes Inception net to extract features from generated images and real ones, and then compares their statistical difference. sFID differs from FID in using intermediate spatial features. IS concentrates more on the fidelity of the images. Precision and Recall evaluates the quality and diversity of generated images respectively. 
% Unless stated otherwise, evaluation results we reported don't use classifier-free guidance. 

\begin{table}[t]
    \caption{Benchmarking unconditional image generation on CelebA-HQ $256\times 256$. } %
    \label{tbl:result_celeba}
    \begin{center}
    \begin{small}
    \begin{tabular}{lccc}
    \toprule
    \multicolumn{4}{l}{\bf{Unconditional CelebA-HQ} $256\times 256$} \\
    \toprule
    Model                                 & FID$\downarrow$   & Precision$\uparrow$ & Recall$\uparrow$ \\
    \midrule
    LSGM~\citep{vahdat2021score}          & 7.22  & -     & - \\
    WaveDiff~\citep{phung2023wavelet}     & 5.94  & -     & 0.37 \\
    LDM-4~\citep{rombach2022high}         & 5.11  & 0.72  & 0.49 \\
    StyleSwin~\citep{zhang2022styleswin}  & 3.25  & -     & - \\
    \textbf{RDM}                          & \bf{3.15} & \bf{0.77} & \bf{0.55} \\
    \bottomrule
    \end{tabular}
    \end{small}
    \end{center}

\end{table}

\begin{table}[t]
    \caption{Benchmarking class-conditional image generation on ImageNet $256\times 256$. } %
    \label{tbl:result_ImageNet}
    \begin{center}
    \begin{small}
    \begin{tabular}{lccccc}
    \toprule
    \multicolumn{6}{l}{\bf{Class-Conditional ImageNet} $256\times 256$} \\
    \toprule
    Model & FID$\downarrow$   & sFID$\downarrow$  & IS$\uparrow$     & Precision$\uparrow$ & Recall$\uparrow$ \\
    \midrule
    BigGAN-deep~\citep{brock2018large} & 6.95 & 7.36 & 171.4 & 0.87 & 0.28 \\
    StyleGAN-XL~\citep{sauer2022stylegan} & 2.30 & 4.02 & 265.12 & 0.78 & 0.53 \\
    \midrule
    ADM~\citep{dhariwal2021diffusion}       & 10.94 & 6.02 & 100.98 & 0.69 & 0.63 \\
    LDM-4~\citep{rombach2022high}           & 10.56 & - & 103.49 & 0.71 & 0.62 \\
    DiT-XL/2~\citep{peebles2022scalable}    & 9.62 & 6.85 & 121.50 & 0.67 & 0.67 \\
    MDT-XL/2~\citep{gao2023masked}          & 6.23 & 5.23 & 143.02 & 0.71 & 0.65 \\
    \textbf{RDM}                            & \bf{5.27} & 4.39 & 153.43 & 0.75 & 0.62 \\
    \cmidrule(lr){1-6}
    CDM~\citep{ho2022cascaded}              & 4.88 & -      & 158.71 & -    & - \\
    ADM-U,G                                 & 3.94 & 6.14   & 215.84 & 0.83 & 0.53 \\
    LDM-4-G (CFG=1.50)                      & 3.60 & -      & 247.67 & 0.87 & 0.48 \\
    \cmidrule(lr){1-6}
    MDT-XL/2-G (dynamic\ CFG) & \textbf{1.79} & 4.57 & 283.01 & 0.81 & 0.61 \\
    \cmidrule(lr){1-6}
    DiT-XL/2-G (CFG=1.50)   & 2.27 & 4.60 & 278.24 & 0.83 & 0.57 \\
    MDT-XL/2-G (CFG=1.325)  & 2.26 & 4.28 & 246.06 & 0.81 & 0.59 \\
    \textbf{RDM} (CFG=3.50) & 1.99 & 3.99 & 260.45 & 0.81 & 0.58 \\
    \quad + class-balance & \textbf{1.87} & \bf{3.97} & 278.75 & 0.81 & 0.59 \\
    \bottomrule
    \end{tabular}
    \end{small}
    \end{center}
\end{table}

% \begin{table}[h]
%     \begin{center}
%     \begin{small}
%     \begin{tabular}{lcc}
%     \toprule
%     Model       & Number of Trained Images   & FID$\downarrow$   \\
%     \midrule
%     DiT-XL/2    & 600M                    & 10.67   \\
%                 & 1792M                    & 9.62    \\
%     \cmidrule(lr){1-3}
%     MDT-XL/2    & 332M                     & 9.60    \\
%                 & 640M                     & 7.41    \\
%                 & 896M                     & 6.46    \\
%                 & 1664M                    & 6.23    \\
%     \cmidrule(lr){1-3}
%     \textbf{RDM}& 365M                     & 8.25    \\
%                 & 536M                     & 7.05    \\
%                 & 628M                     & 6.11    \\
%                 & 791M                     & 5.65    \\
%                 & 1234M                    & 5.27    \\ 
%     \bottomrule
%     \end{tabular}
%     \end{small}
%     \end{center}
%     \caption{Comparison of FID-50K under different training quantities on ImageNet $256\times 256$. The results of DiT-XL/2 and MDT-XL/2 are obtained from their papers respectively~\citep{peebles2022scalable}~\citep{gao2023masked}. } 
%     \label{tbl:ImageNet-cmp}
% \end{table}

\subsection{Results}
\textbf{CelebA-HQ} We compare RDM with the existing methods on CelebA-HQ $256\times 256$ in Table~\ref{tbl:result_celeba}. RDM outperforms the state-of-the-art model StyleSwin~\citep{zhang2022styleswin} with a remarkably fewer training iterations (50M versus 820M trained images). We also achieve the best precision and recall among the existing works.

\textbf{ImageNet} Table~\ref{tbl:result_ImageNet} shows the performance of class-conditional generative models on ImageNet $256\times 256$. We report the best results as possible of the existing methods with classifier-free guidance (CFG)~\citep{ho2022classifier}. RDM achieves the best sFID and outperforms all the other methods by FID except MDT-XL/2~\citep{gao2023masked} with a dynamic CFG scale. 
If with a fixed but best-picked CFG scale\footnote{The best CFG scale is 1.325 with a hyperparameter sweep from 1.0 to 1.8. We observed the FID increases greatly if CFG scale $>$ 1.5 for MDT-XL/2.}, MDT-XL/2 can only achieve an FID of 2.26. 
% For RDM, we disable CFG in the first stage while always use a fixed CFG scale for the second stage.
While achieving competitive results, RDM is trained with only 70\% of the iterations of MDT-XL/2 (1.2B versus 1.7B trained images), indicating that the longer training and a more granular CFG strategy are potential directions to further optimize the FID of RDM.

% This might indicate that a meticulous search of the dynamic CFG scale is an indispensable procedure of the best FID, which is not the focus of our work. For RDM, we disable CFG in the generation of the first stage while always use a fixed CFG scale for the second stage.

\textbf{Training Efficiency} We also compare the performance of RDM with existing methods along with the training cost in Figure~\ref{fig: main_results}. When CFG is disabled, RDM achieves a better FID than previous state-of-the-art diffusion models including DiT~\citep{peebles2022scalable} and MDT~\citep{gao2023masked}. RDM outperforms them even with only about $1/3$ training iterations. 

% RDM demonstrate a faster convergence compared to the SoTA model. For ImageNet $256\times 256$, as shown in Figure \ref{fig: main_results}, RDM achieving a nearly three-fold increase in learning speed compared to MDT and DiT. RDM also leads to a significant improvement in FID compared to MDT, from 6.23 to 5.27.
% For CelebA-HQ $256\times 256$, as shown in Table \ref{tbl:result_celeba}, RDM outperforms the SoTA model StyleSwin~\citep{zhang2022styleswin}, even though StyleSwin is trained on 820M images, tens of times more than the 50M images used for training RDM.
% RDM achieves better performance with a faster learning speed, as it utilizes a 64-resolution model in the first stage and subsequently only needs to be trained on a truncated noise schedule in the second stage.

% When using CFG on ImageNet $256\times 256$, RDM achieved a competitive result compared to the SoTA model MDT, even though it was trained on less than half of the images used by MDT. Moreover, RDM obtain the best sFID and achieved an FID of 1.97, outperforming all previous diffusion models using fixed CFG scale (when MDT uses a fixed CFG scale, its best performance can only reach an FID of 2.26 after iterating through the values of CFG scale). 
% MDT utilizes a meticulously designed dynamic CFG scale with a power-cosine schedule. Since this is not the focus of our work, we used a fixed CFG scale in the second stage and do not use CFG in the first stage.

\subsection{Ablation Study}\label{sec:ablation}

In this section, we conduct ablation experiments on the designs of RDM to verify their effectiveness. Unless otherwise stated, we report results of RDM on $256\times 256$ generation without CFG.

\begin{figure}[h]
\begin{center}
{\includegraphics[width=1.0\linewidth]{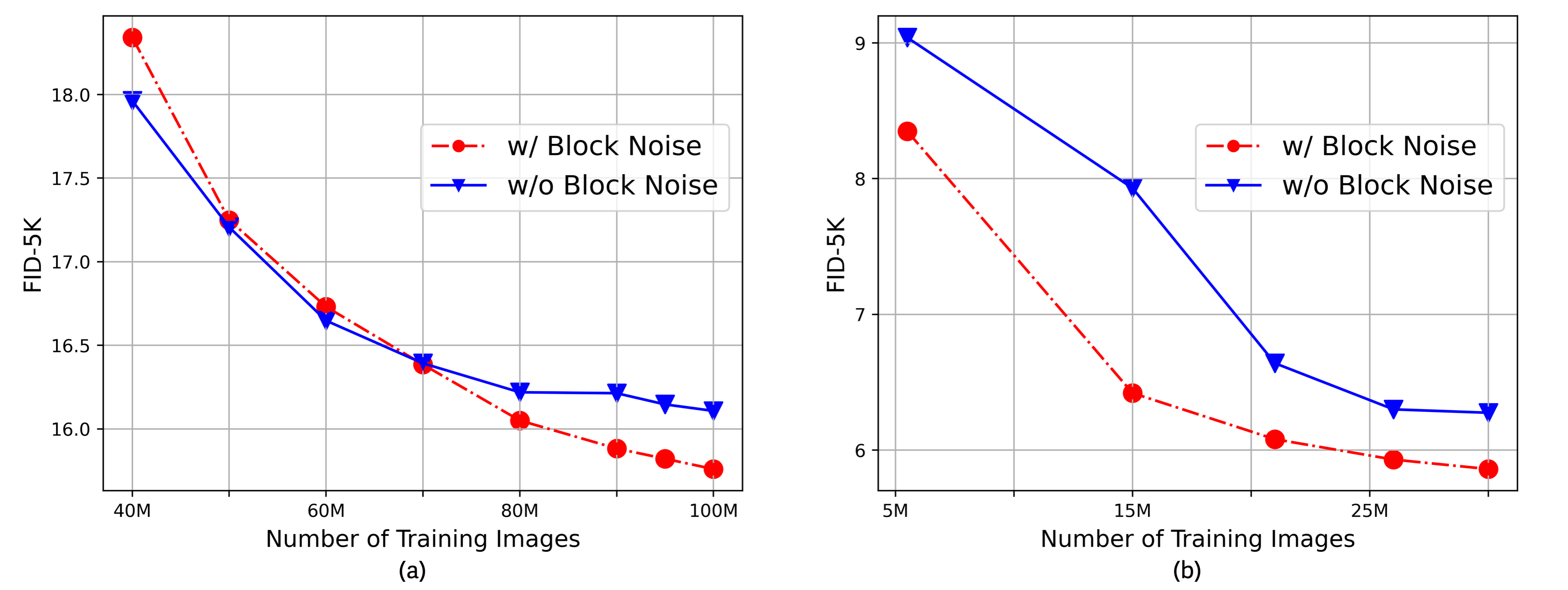}}
\vspace{-7mm}
\end{center}
\caption{The effectiveness of block noise. We compare the performance of RDM along the training on (a) ImageNet $256\times 256$ and (b) CelebA-HQ $256\times 256$. To apply block noise in RDM, we set $\alpha=0.15$ and kernel size $s=4$.}
\label{fig: ablation_block}
\end{figure}

\textbf{The Effectiveness of block noise.} \label{sec:ablation_blocknoise}
We compare the performance of RDM with and without adding block noise in Figure~\ref{fig: ablation_block}. With a sufficient phase of training, RDM with block noise outperforms the model without block noise by a remarkable margin on both ImageNet and CelebA-HQ. This demonstrates the effectiveness of the block noise. The addition of block noise introduces higher modeling complexity of the noise pattern, which contributes to a slower convergence of training in the initial stage, as illustrated by Figure~\ref{fig: ablation_block}(a). We assume that training on a significantly smaller scale of samples leads to a fast convergence of the model, which obliterates such a feature, therefore a similar phenomenon cannot be observed in the training of CelebA-HQ.

\begin{table}[h]
    \centering
    \begin{tabular}{lcccccccccc}
    \toprule
    $\eta$              & 0    & 0.10 & 0.15 & 0.20      & 0.25 & 0.30 & 0.40 & 0.50 \\
    \midrule
    FID$\downarrow$     & 5.65 & 5.44 & 5.31 & \bf{5.27} & 5.48 & 5.91 & 6.91 & 9.17 \\
    \bottomrule
    \end{tabular}
    
    \vspace{15pt}  % 调整两个表格之间的垂直间距
    
    \begin{tabular}{lcccccccccc}
    \toprule
    $\eta$              & 0    & 0.10 & 0.15 & 0.20      & 0.25 & 0.30 & 0.40 & 0.50 \\
    \midrule
    FID$\downarrow$     & 4.11 & 3.74 & 3.43 & \bf{3.15} & 3.23 & 3.52 & 4.79 & 6.41 \\
    \bottomrule
    \end{tabular}
    \captionof{table}{ Effect of stochasticity in the sampler on ImageNet $256\times 256$ (top) and CelebA-HQ $256\times 256$ (bottom). We explored different values of the $\eta$ in Algorithm \ref{alg:sample}.} 
    \label{tbl:ablation_sde}
\end{table}

\textbf{The scale of stochasticity.} 
As previous works~\citep{song2020score} have shown, SDE samplers usually perform better than ODE samplers. 
% To improve the quality of generation, we design a stochastic sampler for RDM to inject noise into the process of sampling. 
We want to quantitatively measure how the scale of the stochaticity affects the performance in the RDM sampler (Algorithm~\ref{alg:sample}). Table~\ref{tbl:ablation_sde} shows results with $\eta$ varying from $0$ to $0.50$. For both CelebA-HQ and ImageNet, the optimal FID is achieved by $\eta=0.2$. We hypothesis a small $\eta$ is insufficient for the noise addition to cover the bias formed in earlier sampling steps, while a large $\eta$ introduces excessive noise into the process of sampling, which makes a moderate $\eta$ to be the best choice. Within a reasonable scale of stochasticity, an SDE sampler always outperforms the ODE sampler by a significant margin.

% \textbf{SDE Sampler.} \label{sec:ablation_sdesampler}
% As mentioned in Section \ref{sec:sampler}, we design a stochastic sampler that exhibits second order convergence. 
% We compare the effects of different levels of stochasticity on image generation in Table \ref{tbl:ablation_sde}. 
% The optimal choice for $\eta$ is 0.2. If $\eta$ is too small, the provided stochasticity is insufficient to correct the biases introduced in the early sampling steps, resulting in a near degeneracy into an ODE sampler. Conversely, if $\eta$ is too large, it introduces excessive noise, presenting additional challenges for the denoising process. Both scenarios above lead to a deterioration in performance.
% It is indicated that the SDE sampler, with an appropriate selection of $\eta$, outperforms the ODE sampler ($\eta=0$).

\begin{wrapfigure}[23]{r}{.55\textwidth}
\vspace{-4mm}
    \centering
\includegraphics[width=0.55\textwidth]{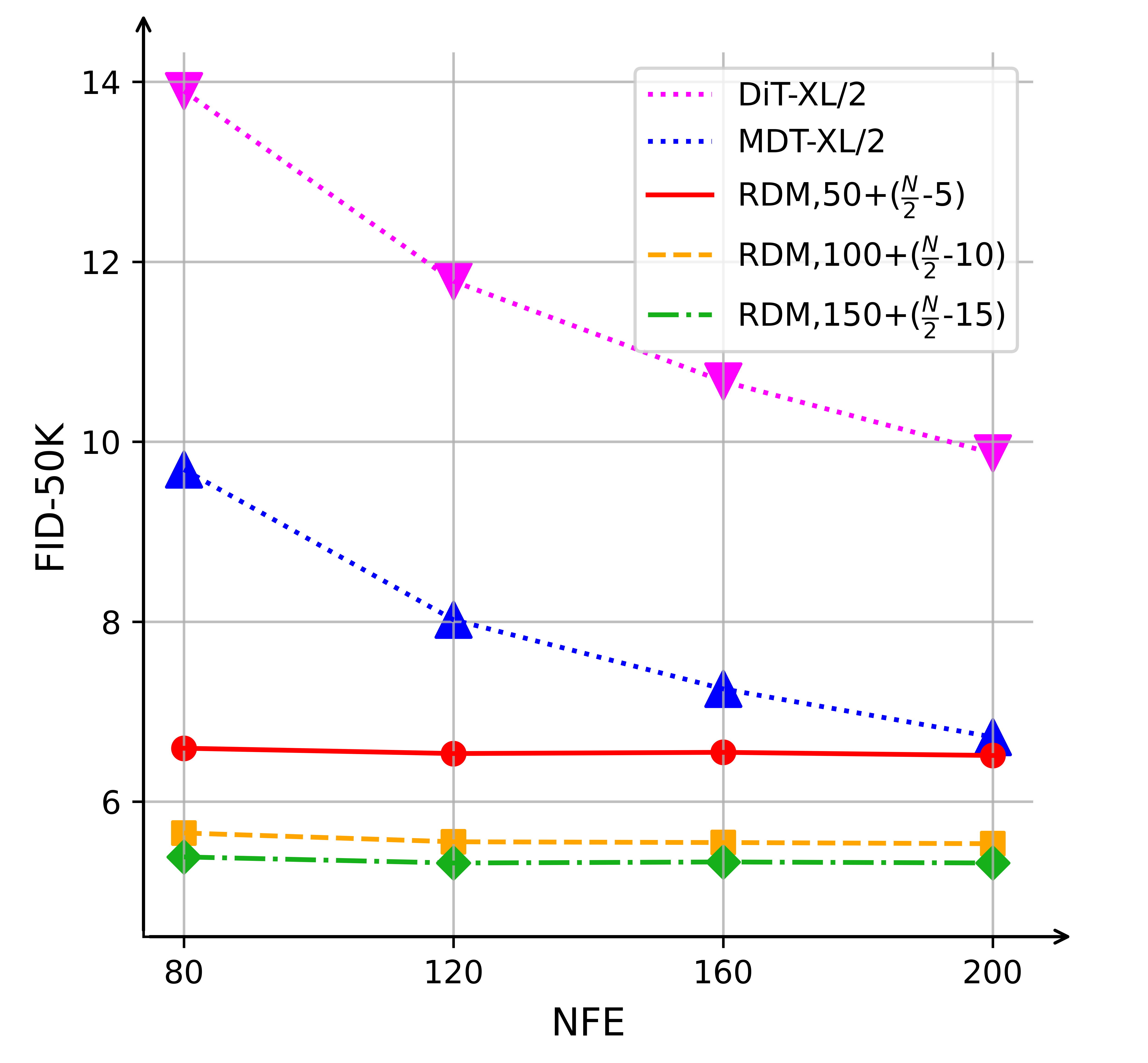}
\vspace{-4mm}
    \caption{Comparison of FID on ImageNet with different sampling steps. For allocation of NFE $=N$ in RDM, $10n+(\frac{N}{2}-n)$ means $10n$ for the first stage and $\frac{N}{2}-n$ for the second.}
    \label{fig:ablation_steps}
\end{wrapfigure}

\textbf{Sampling steps.} To demonstrate the efficiency of our model, we compare the performance of RDM and other methods with fewer sampling steps. Number of Function Evaluations (NFE), i.e., the number that a neural network is called during sampling, is used as the index of the comparison for fairness. For RDM, the NFE consists of the NFE in the second stage and $1/10$ the NFE in the first stage, according to the proportion of the FLOPs. As shown in Figure~\ref{fig:ablation_steps}, the performance of DiT-XL/2~\citep{peebles2022scalable} and MDT-XL/2~\citep{gao2023masked} both drop significantly with a lower NFE, while RDM barely declines. Considering that the steps in different stages may contribute differently in FID, we demonstrates three FLOPs allocation strategies in Figure~\ref{fig:ablation_steps}.  
% Ablation is also conducted to compare settings of RDM with varied distribution of NFE between the two stages. 
With more NFE allocated in the first stage, RDM achieves a better FID. In all the settings, RDM performs better than MDT-XL/2 and DiT-XL/2 if NFE $<$ 200.
%  which indicates that a sufficient NFE allocation in the first stage ensures a good starting point of the second stage generation, and therefore improves the final performance.

% \textbf{Sampling Steps.}
% We use number of function evaluations (NFE) to uniformly compare the sampling steps of different models. The NFE of the first stage of RDM can be equivalently represented as one-tenth of the NFE of the second stage, based on their respective GFLOPs. 
% We compare the performance of RDM, MDT, and DiT under different NFE in Figure \ref{fig:ablation_steps}, where we fix the NFE for the first stage of RDM at 200, equivalent to 20 NFE for the second stage.
% When NFE is low, RDM demonstrates significant advantages over MDT and DiT. This is attributed to the fact that (1) RDM starts diffusion from the low-resolution result instead of pure noise, significantly reducing the required NFE for generating high-quality samples.
% (2) Our second order sampler accelerates the sampling process.

% \begin{figure}[h]
% \begin{center}
% {\includegraphics[width=1.0\linewidth]{figures/ablation_steps.png}}
% \vspace{-7mm}
% \end{center}
% \caption{Comparison of different NFE on ImageNet $256\times 256$.}
% \label{fig: ablation_steps}
% \end{figure}

\section{Conclusion and Discussion}

% In this work, we propose Relay Diffusion Model (RDM) to handle diffusion generation, which disentangles the diffusion process into a cascaded pipeline and introduces the block noise to bridge the gap between generations of different resolutions. Compared to previous cascaded diffusion models, each stage of RDM starts generation from results of the last stage. The design of RDM better aligns with the nature of images and reduces the cost of generating high-resolution samples.
% Our experiments show that RDM achieves a competitive performance while greatly reducing the consumption of training as well as inference. When classifier-free guidance (CFG) is disabled, our model obtains a better quality of generation than previous SoTA diffusion models.

In this paper, we propose relay diffusion to optimize the cascaded pipeline. The diffusion process can now continue when changing the resolution or model architectures. We anticipate that our method can reduce the cost of training and inference, and help create more advanced text-to-image model in the future.

The frequency analysis in the paper reveals the relation between noise and image resolution, which might be helpful to design a better noise schedule. 
However, our numerous attempts to directly derive the optimal noise schedule on the dataset did not yield good results. The reason might be that the optimal noise schedule is also be related to the size of the model, inductive bias, and the nuanced distribution characteristics of the data. Further investigation is left for future work.

\subsubsection*{Author Contributions}
Ming Ding proposes the methods and leads the project. Jiayan Teng and Wendi Zheng conduct most of the experiments. Wenyi Hong works together on early experiments.  Jianqiao Wangni, Wenyi Hong and Zhuoyi Yang contribute to the writing of the paper. Jie Tang provides guidance and supervision.

The work is partly done during the internship of Jiayan Teng and Wendi Zheng at Zhipu AI.
\subsubsection*{Acknowledgments}
The authors also thank Ting Chen from Google DeepMind and Junbo Zhao from Zhejiang University for their valuable talks and comments.

\bibliography{iclr2023_conference}
\bibliographystyle{iclr2023_conference}

\newpage
\appendix

\section{Derivation}

\subsection{Patch-wise Blurring}\label{sec:patch_blur}
% \tjy{version-2}

The forward process of blurring diffusion is defined as Eq.~\ref{eq:blur-diff}, where $\vu_0=\mV^\mathrm{T} \vx_0$ denotes the representation of the image $\vx_0$ in the frequency space. The diagonal matrix $\mD_t=e^{\mathbf{\Lambda} t}$ defines a non-isotropic blurring projection, where $\mathbf{\Lambda}(i\times W+j,i\times W+j)=-\pi^2(\frac{i^2}{H^2}+\frac{j^2}{W^2})$ corresponds to the coordinate $(i, j)$ in the 2D frequency space. In the equation $q(\vu_t | \vu_0) = \mathcal{N}(\vu_t | \mD_t \vu_{0}, \sigma_t^2\mI)$, we can utilize the dot product of matrices to transform $\mD_t$ and $\vu_0$ into 2D matrices, $\Tilde{\mD_t}$ and $\Tilde{\vu_0}$, in the shape of $H\times W$ for calculation:
\begin{equation}
    \mD_t \vu_0 \Rightarrow \Tilde{\mD_t} \cdot \Tilde{\vu_0}
\end{equation}

In the super-resolution stage of RDM, we apply blurring on each $k\times k$ patch independently. We name it as patch-wise blurring and define the diagonal blurring matrix in the shape of $k\times k$ for each patch as:
\begin{equation}
    \Tilde{\mD}_{t,k\times k}=\exp(\Tilde{\mathbf{\Lambda}}_{k\times k} t),\quad \Tilde{\mathbf{\Lambda}}_{k\times k}(i,j)=-\pi^2(\frac{i^2}{k^2}+\frac{j^2}{k^2}),
\end{equation}
where $i\in [0,k), j\in [0,k)$. For any patch, $\Tilde{\mD}_{t,k\times k}$ remains the same. The blurring matrix $\Tilde{\mD}^p_t$ of the patch-wise blurring is a combination of all the independent blurring matrices $\Tilde{\mD}_{t,k\times k}$. The relationship between the elements of $\Tilde{\mD}^p_t$ and $\Tilde{\mD}_{t,k\times k}$ can be expressed as:
\begin{equation}
    \Tilde{\mD}^p_t(i,j)=\Tilde{\mD}_{t,k\times k}(i\ \text{mod}\ k, j\ \text{mod}\ k),
\end{equation}
where $(i,j)$ corresponds to the coordinate in the 2D frequency space. Finally, $\mD^p_t$ in Eq.~\ref{eq:patch-blur} can be formulated as:
\begin{equation}
    \mD^p_t=\text{diag}(\text{unfold}(\Tilde{\mD}^p_t)),
\end{equation}
where $\text{unfold}(\Tilde{\mD}^p_t)$ means unfolding the $H\times W$ matrix into  a vector of $HW$ dimensions and $\text{diag}(\vv)$ denotes the diagonal matrix with vector $\vv$ as its diagonal line.

\subsection{Combination of Schedule}

We follow~\cite{karras2022elucidating} to set the noise schedule for standard diffusion as $\ln(\sigma)\sim\mathcal{N}(P_{mean}, P^2_{std})$. We use $\mathcal{F}_{\mathcal{D}}$ and $\mathcal{F}^{-1}_{\mathcal{D}}$ to denote the cumulative distribution function (CDF) and the inverse distribution function (IDF) for distribution $\mathcal{D}$ in the following description. With $t$ sampled from uniform distribution $\mathcal{U}(0,1)$, the noise scale is formulated as:
\begin{equation}
    \sigma(t)=\exp(\mathcal{F}^{-1}_{\mathcal{N}(P_{mean}, P^2_{std})}(t)).
\end{equation}

For the super-resolution stage of RDM, we apply a truncated version of diffusion noise schedule $\sigma'(t), t\sim\mathcal{U}(0,1)$. If we set $t_s$ as the starting point of the truncation, the new noise schedule can be formally expressed as:
\begin{equation}
    \sigma'(t)=\sigma(\mathcal{F}^{-1}_{\mathcal{U}(0,1)}(\mathcal{F}_{\mathcal{U}(0,1)}(t_s)\mathcal{F}_{\mathcal{U}(0,1)}(t))),
\end{equation}
which means we only sample the noise scale $\sigma'$ from positions of the normal distribution $\mathcal{N}(P_{mean}, P^2_{std})$ where its CDF is less than $t_s$.

For the process of blurring, we set its schedule following the setting of \cite{hoogeboom2022blurring}. They found that the heat dissipation is equivalent to a Gaussian blur with the variance of its kernel as $\sigma^2_{B,t}=2\tau_t$. They set the blurring scale $\sigma_{B,t}$ as:
\begin{equation}
    \sigma_{B,t}=\sigma_{B,max}\sin^2(\frac{t\pi}{2}),
\end{equation}
where $t$ is also sampled from the uniform distribution $\mathcal{U}(0,1)$ and $\sigma_{B,max}$ denotes a fixed hyperparameter. Empirically, we set $\sigma_{B,max}=3$ for ImageNet $256\times 256$ and $\sigma_{B,max}=2$ for CelebA-HQ $256\times 256$. The blurring matrix is formulated as $\mD_t=e^{\bm{\Lambda}\tau_t}$, where $\tau_t=\frac{\sigma^2_{B,t}}{2}$. As illustrated in Section~\ref{sec:ihd}, $\mathbf{\Lambda}$ is a diagonal matrix and $\mathbf{\Lambda}_{i\times W+j}=-\pi^2(\frac{i^2}{H^2}+\frac{j^2}{W^2})$ for coordinate $(i,j)$.

\subsection{Sampler Derivation}\label{sec:sampler_deriv}
In this section, we prove the consistency between the design of our sampler and the formulation of blurring diffusion. We need to prove that the jointly distribution $q_{\delta}(\vu_{n-1}|\vu_n,\vu_0)$ we define in Eq.~\ref{eq:sampler_dist} matches with the marginal distribution 
\begin{equation}
    q_{\delta}(\vu_n | \vu_0) = \mathcal{N}(\vu_n | \mD^p_{t_n} \vu_{0},\sigma_{t_n}^2\mI)
\end{equation}
under the condition of Eq.~\ref{eq:family_dist}.

\textit{proof.}\  Given that $q_{\delta}(\vu_N | \vu_0) = \mathcal{N}(\vu_N | \mD^p_{t_N} \vu_{0},\sigma_{t_N}^2\mI)$, we proceed with a mathematical induction approach. Assuming that for any $n \leq N$, $q_{\delta}(\vu_n | \vu_0) = \mathcal{N}(\vu_n | \mD^p_{t_n} \vu_{0},\sigma_{t_n}^2\mI)$ holds. We only need to prove $q_{\delta}(\vu_{n-1} | \vu_0) = \mathcal{N}(\vu_{n-1} | \mD^p_{t_{n-1}} \vu_{0},\sigma_{t_{n-1}}^2\mI)$, and then the conclusion above will be proved based on the induction hypothesis.

Firstly, based on
\begin{equation} \label{eq:jointly_dist}
    q_{\delta}(\vu_{n-1} | \vu_0) = \int q_{\delta}(\vu_{n-1} | \vu_n, \vu_0)q(\vu_n | \vu_0)d\vu_n,
\end{equation}
we introduce
\begin{equation}
    q_{\delta}(\vu_{n-1}|\vu_n,\vu_0)=\mathcal{N} \big(\vu_{n-1}|\frac{1}{\sigma_{t_n}}(\sqrt{\sigma^2_{t_{n-1}}-\delta^2_n}\vu_n+(\sigma_{t_n}\mD^p_{t_{n-1}}-\sqrt{\sigma^2_{t_{n-1}}-\delta^2_n}\mD^p_{t_n})\vu_0),\delta^2_n\mI \big)
\end{equation}
and
\begin{equation}
    q_{\delta}(\vu_n | \vu_0) = \mathcal{N}(\vu_n | \mD^p_{t_n} \vu_{0},\sigma_{t_n}^2\mI).
\end{equation}
Then according to~\citet{bishop2006pattern}, $q_{\delta}(\vu_{n-1} | \vu_0)$ is also a Gaussian distribution:
\begin{equation}
    q_{\delta}(\vu_n | \vu_0) = \mathcal{N}(\vu_n | \bm{\mu}_{n-1}, {\bf{\Sigma}}_{n-1}).
\end{equation}
Therefore, from Eq.~\ref{eq:jointly_dist}, we can derive that
\begin{equation}
    \bm{\mu}_{n-1} 
    = \frac{1}{\sigma_{t_n}}(\sqrt{\sigma^2_{t_{n-1}}-\delta^2_n}\mD^p_{t_n} \vu_{0}+(\sigma_{t_n}\mD^p_{t_{n-1}}-\sqrt{\sigma^2_{t_{n-1}}-\delta^2_n}\mD^p_{t_n})\vu_0)
     = \mD^p_{t_{n-1}} \vu_{0}
\end{equation}
and
\begin{equation}
    {\bf{\Sigma}}_{n-1} = \frac{\sigma^2_{t_{n-1}}-\delta^2_n}{\sigma^2_{t_n}}\sigma^2_{t_n}\mI + \delta^2_n\mI  = \sigma^2_{t_{n-1}}\mI.
\end{equation}
Summing up, $q_{\delta}(\vu_{n-1} | \vu_0) = \mathcal{N}(\vu_{n-1} | \mD^p_{t_{n-1}} \vu_{0},\sigma_{t_{n-1}}^2\mI)$. The inductive proof is complete.

\section{Hyperparameters}

Hyperparameters we use for the training of RDM are presented in Table~\ref{tbl:hypers}. We set the architecture hyperparameters for diffusion models following~\cite{dhariwal2021diffusion}, in corresponding to the input resolutions. For the experiments on CelebA-HQ, we set the model dropout to be larger (0.15 and 0.2 for two stages respectively), and enable sample augmentation to prevent RDM from overfitting.

\label{app:hypers}
\begin{table}[h]
    \setlength\tabcolsep{4pt}
    \begin{center}
    \begin{tabular}{lcccc}
    \toprule
     & ImageNet 64 & ImageNet 64$\rightarrow$256 & CelebA-HQ 64 & CelebA-HQ 64$\rightarrow$256 \\
    \midrule
    Diffusion steps & 256 & 100 & 120 & 53 \\
    Noise Schedule & cosine & linear & linear & linear \\
    Model size  & 295M & 553M & 295M & 553M \\
    GFLOPs      & 104 & 1117 & 104 & 1117 \\
    Mixed-precision (FP16) & \checkmark & \checkmark & - & \checkmark \\
    Channels & 192 & 256 & 192 & 256  \\
    Channels multiple & 1,2,3,4 & 1,1,2,2,4,4 & 1,2,3,4 & 1,1,2,2,4,4 \\
    Heads Channels & 64 & 64 & 64 & 64 \\
    Attention resolution & 32,16,8 & 32,16,8 & 32,16,8 & 32,16,8 \\
    Dropout & 0.1 & 0.1 & 0.15 & 0.2 \\
    Augment probability & 0 & 0 & 0.2 & 0.2 \\
    Blurring $\sigma_{max}$ & - & 3.0 & - & 2.0 \\
    Batch size & 4096 & 4096 & 1024 & 1024 \\
    Training Images & 2500M & 1000M & 70M & 40M \\
    Learning Rate & 1e-4 & 1e-4 & 1e-4 & 1e-4 \\
    \bottomrule
    \end{tabular}
    \end{center}
    \caption{Hyperparameters for RDM.}
    \label{tbl:hypers}
     \vskip -0.2in
\end{table}

\section{Details About The Power Spectral Density}
\label{app:psd}

\subsection{Calculation Procedure of the PSD}
We follow the setting of~\citet{rissanen2022generative} to calculate the PSD in the frequency space. The PSD at a certain frequency is defined as the square of the DCT coefficient at that frequency. Firstly, we transform the image into the 2D frequency space by DCT and set the frequency range to [0, $\pi$]. To obtain the 1D curve of the PSD, we calculate the distance from each point ($x$, $y$) to the origin in the frequency space, i.e. $\sqrt{x^2+y^2}$, considering it as a 1D frequency value. Subsequently, we uniformly divide the frequency domain into $N$ intervals, and take the midpoint of each interval as its representative frequency value. Finally, we take the mean of the PSD values for all points within the interval as the PSD value for that interval, in order to get $N$ coordinate pairs for plotting. The SNR curve in Figure~\ref{fig: block} can be obtained in a similar approach, while the only difference is that the vertical axis values are replaced with the absolute value of the ratio between the DCT coefficients for the image and noise in the frequency space.

\subsection{Analysis of the PSD}
\begin{figure}[h]
\begin{center}
{\includegraphics[width=1.0\linewidth]{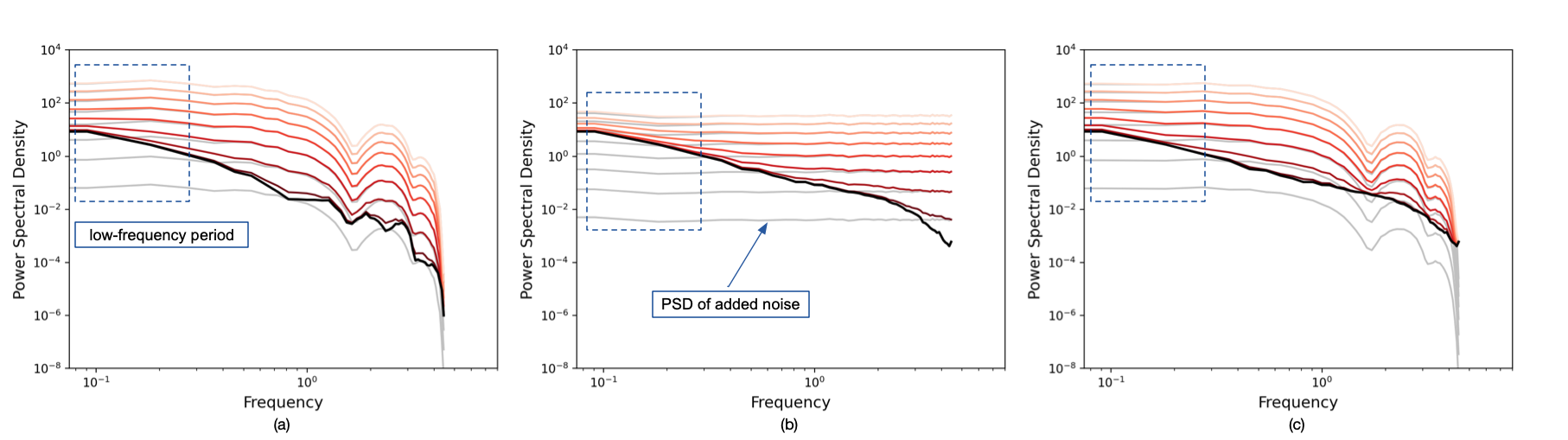}}
\vspace{-7mm}
\end{center}
\caption{The power spectral density (PSD) of real images after adding (a) 64px Gaussian noise, (b) 256px Gaussian noise and (c) 256px block noise with block size of 4. The black curve represents the PSD of real images. The red curves, from dark to light, represent adding noise with increasing intensity. In order to make comparisons within the same frequency space, for the images at the resolution of 64, we firstly upsample them to the pixel space at the resolution of 256.}
\label{fig:psd}
\end{figure}

As shown in Figure~\ref{fig:psd}, the PSD of real images gradually decreases from low frequency to high frequency. And the intensity of Gaussian noise components across all frequency bands is generally equal. Therefore, when corrupting real images, Gaussian noise initially drowns out high-frequency components until the noise intensity becomes high enough to drown out the low-frequency components of real images. And it is demonstrated in Figure~\ref{fig: block} that, as the resolution of images increases, less information is corrupted under the same noise intensity. Correspondingly, as shown in Figure~\ref{fig:psd}(a) and Figure~\ref{fig:psd}(b), the low-frequency portion of the PSD gets drowned out more slowly as the resolution increases. It is indicated that we will introduce excessive high-frequency components of noise when corrupting the low-frequency information of real images, especially for high-resolution images. 

Differently, the low-frequency portion of the PSD from block noise is notably higher than that of Gaussian noise with the same intensity. Furthermore, the PSD of block noise exhibits a decreasing trend as frequency increases, and its curve is quite similar to the PSD curve of Gaussian noise at the resolution of 64 upsampled to the resolution of 256. This leads to the PSD curves of high-resolution images with added block noise and that of low-resolution images with added Gaussian noise also being quite similar. As a result, the low-frequency portion of the PSD from images with added block noise gets drowned out more quickly than that from images with added Gaussian noise. We can conclude that block noise can corrupt the low-frequency components of images more easily.

\section{Additional Samples}

\begin{figure}[h]
\begin{center}
{\includegraphics[width=1.0\linewidth]{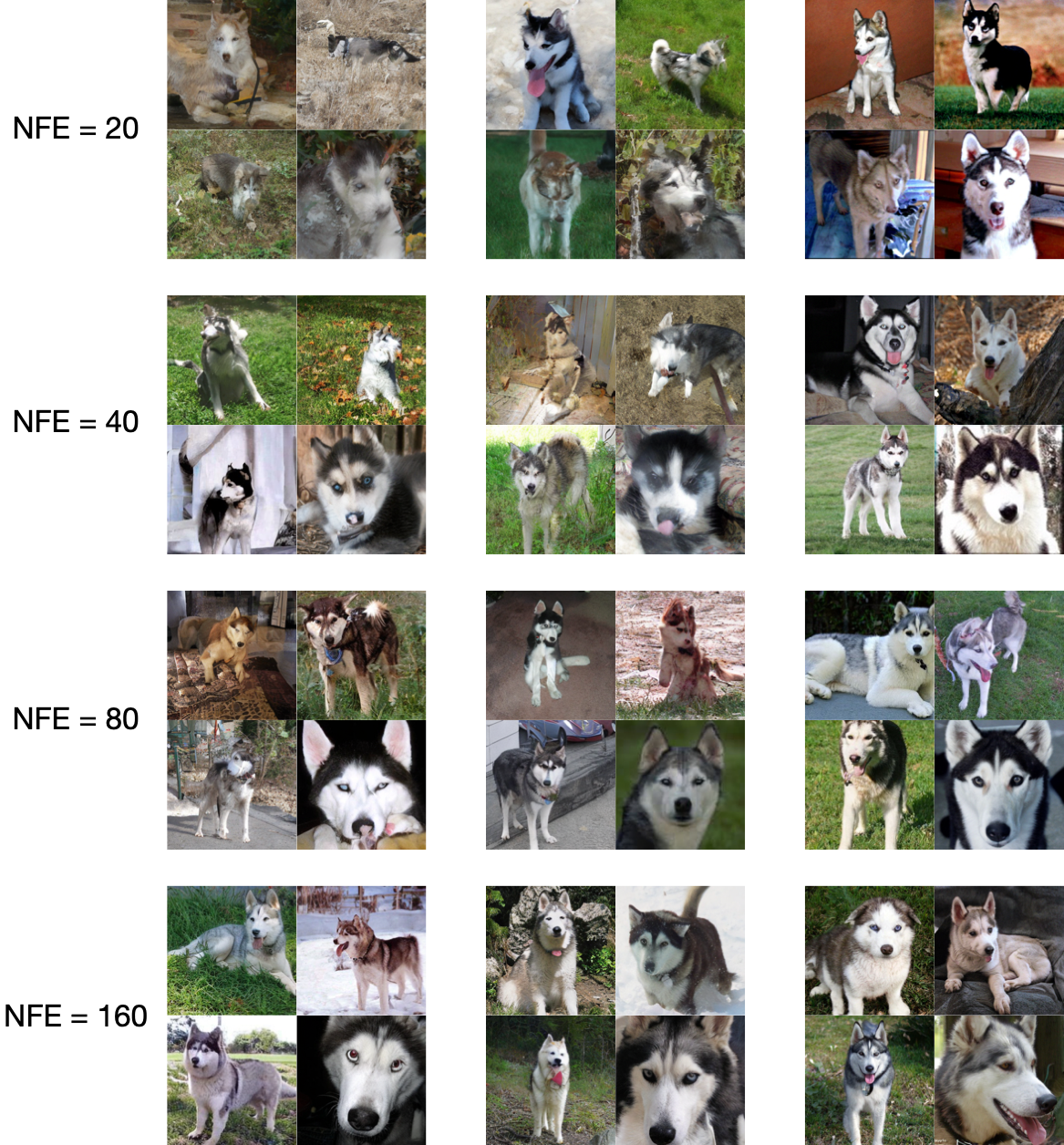}}
\vspace{-7mm}
\end{center}
\caption{Comparison of ImageNet samples with varied NFE. DiT-XL/2 (left) vs MDT-XL/2 (middle) vs RDM (right). The allocation of NFE between the two stages of RDM is: [2, 18], [8, 32], [20, 60], [40, 120]. }
\label{fig:diff_nfe}
\end{figure}

Section~\ref{sec:ablation} quantitatively compares the performance of RDM with other models under the same NFE and demonstrates the superiority of RDM with fewer sampling steps. Figure~\ref{fig:diff_nfe} shows qualitative comparison results. While other models achieve competitive quality of generation with sufficient NFE, their performances degenerate noticeably with the decrease of NFE. In contrast, RDM still maintains comparable generation quality with a low NFE.

\begin{figure}[h]
\begin{center}
{\includegraphics[width=1.0\linewidth]{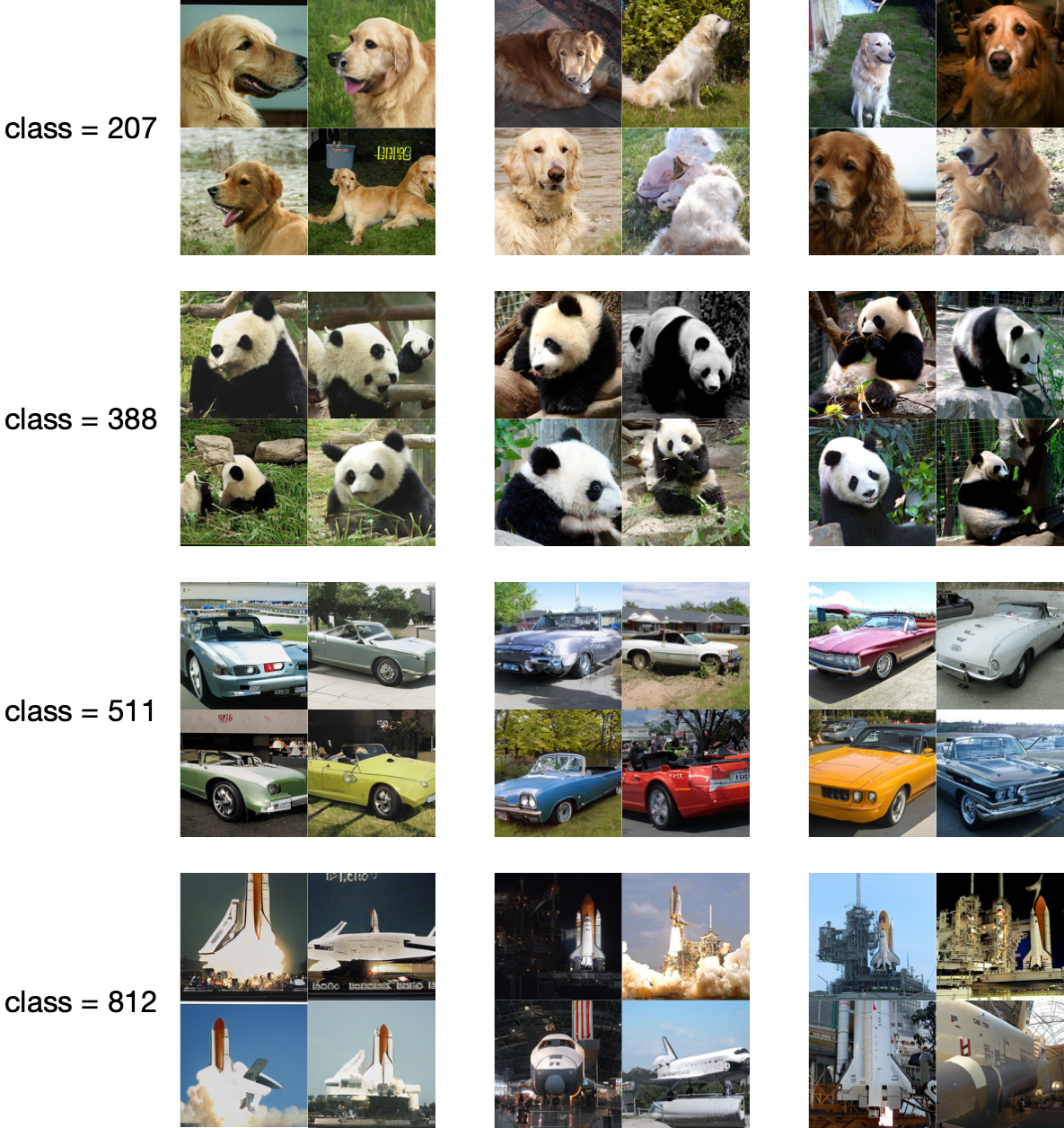}}
\end{center}
\caption{Comparison of best ImageNet samples. StyleGAN-XL (FID 2.30, left) vs DiT-XL/2 (FID 2.27, middle) vs RDM (FID 1.87, right).}
\label{fig:ablation_samples}
\end{figure}

\begin{figure}[h]
\begin{center}
{\includegraphics[width=1.0\linewidth]{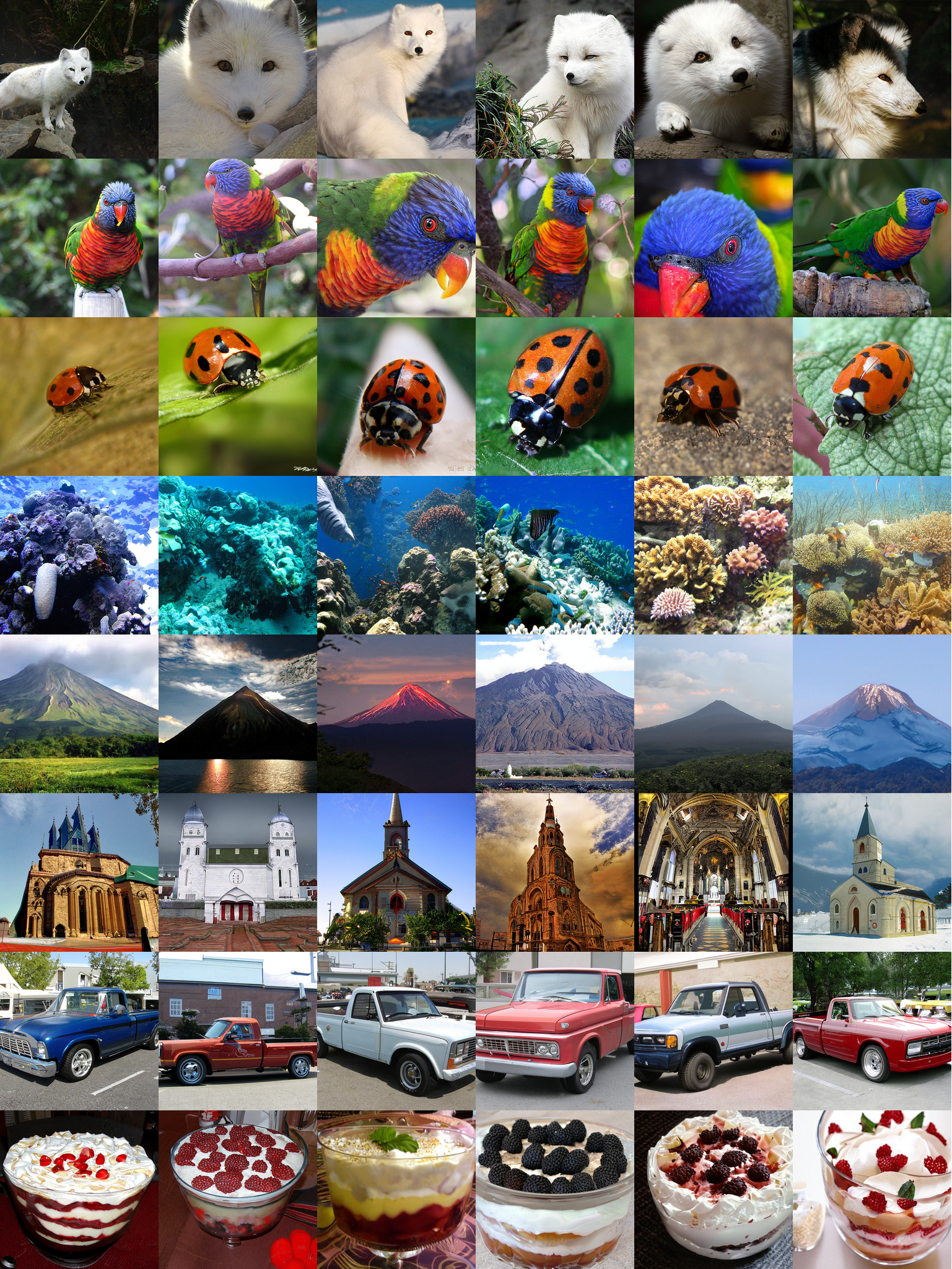}}
\end{center}
\caption{Additional ImageNet samples generated by RDM. Classes are 279: Arctic fox, 90: lorikeet, 301: ladybug, 973: coral reef, 980: volcano, 497: church, 717: pickup truck, 927:  trifle.}
\label{fig:compare_samples}
\end{figure}

Figure~\ref{fig:ablation_samples} compares visualized samples generated by the best settings of StyleGAN-XL~\citep{sauer2022stylegan}, DiT~\citep{peebles2022scalable} and RDM. StyleGAN-XL is in the framework of GAN, while DiT and RDM are diffusion models. RDM achieves the best quality of images synthesis. Figure~\ref{fig:compare_samples} exhibits more examples generated by our model RDM on ImageNet $256\times 256$.

\end{document}